\documentclass[conference]{IEEEtran}
\IEEEoverridecommandlockouts
\usepackage{cite}
\usepackage{amsmath,amssymb,amsfonts}
\usepackage{algorithmic}
\usepackage{graphicx}
\usepackage{textcomp}
\usepackage{xcolor}
\usepackage{comment}
\usepackage{caption}
\usepackage{makecell} 
\usepackage{multirow}
\usepackage{todonotes}
\usepackage{url}            
\usepackage{flushend}

\usepackage{tikz} 
\usepackage{pifont}

\def\BibTeX{{\rm B\kern-.05em{\sc i\kern-.025em b}\kern-.08em
    T\kern-.1667em\lower.7ex\hbox{E}\kern-.125emX}}

\begin{document}

\title{Hyperdimensional Computing  for \\ Efficient Distributed Classification \\ with  Randomized Neural Networks
\thanks{
The work of DK was supported by the European Union's Horizon 2020 Research and Innovation Programme under the Marie Skłodowska-Curie Individual Fellowship Grant Agreement 839179 and in part by the DARPA's VIP (Super-HD Project) and AIE (HyDDENN Project) programs.
}
}
\author{\IEEEauthorblockN{Antonello Rosato}
\IEEEauthorblockA{\textit{DIET Department} \\
\textit{ University of Rome}\\
\textit{``La Sapienza''}\\
Rome, Italy \\
antonello.rosato@uniroma1.it}
\and
\IEEEauthorblockN{Massimo Panella}
\IEEEauthorblockA{\textit{DIET Department} \\
\textit{ University of Rome}\\
\textit{``La Sapienza''}\\
Rome, Italy \\
massimo.panella@uniroma1.it}
\and
\IEEEauthorblockN{Denis Kleyko}
\IEEEauthorblockA{
\textit{UC Berkeley} \\
Berkeley, USA \\
\textit{Research Institutes of Sweden}\\
Kista, Sweden \\
denkle@berkeley.edu}
}

\maketitle

\begin{abstract}

In the supervised learning domain, considering the recent prevalence of algorithms with high computational cost, the attention is steering towards simpler, lighter, and less computationally extensive training and inference approaches.
In particular, randomized algorithms are currently having a resurgence, given their generalized elementary approach.
By using randomized neural networks, we study distributed classification, which can be employed in situations were data cannot be stored at a central location nor shared. 
We propose a more efficient solution for distributed classification by making use of a lossy compression approach applied when sharing the local classifiers with other agents. 
This approach originates from the framework of hyperdimensional computing, and is adapted herein. 
The results of experiments on a collection of datasets demonstrate that the proposed approach has usually higher accuracy than local classifiers and getting close to the benchmark -- the centralized classifier.
This work can be considered as the first step towards analyzing the variegated horizon of distributed randomized  neural networks.
\end{abstract}


\section{Introduction}
\label{sec:intro}

In recent times, randomization techniques in neural networks have been the target of advanced studies, broadening the design prospects and generalization capabilities of different models~\cite{cao2018review}. 
A practical advantage of these  techniques is that they can simplify the training process, which limits required computational costs, and, hence, is an appealing feature in applications where resources are constrained. 

Theoretically speaking, the study of randomized neural networks can enhance the analysis of the inner workings of neural models~\cite{gallicchio2017randomized}, deepening the understanding of certain properties such as interpreting representation inside networks, mapping specific activities to a certain portion of the network or even gain insights on the internal structure and patterns of the weight set. 
While some of the most recent literature~\cite{MONTAVON20181}, \cite{gallicchio2020deep} is steering towards deep randomized neural networks, it is still of great interest to study shallow models for a variety of reasons. 
The trade-off between the accuracy and the efficiency of such models is so delicate that it is far from being considered a closed problem. 
Moreover, there are specific application areas (e.g., low-power, edge, and green computing) in which properties of randomized neural networks make them the sole candidate for solving supervised and unsupervised problems.

Most of the studies of randomized neural networks are done in the centralized scenario where a single agent have access to the full dataset. In fact, the common case study is to analyse the performance of a model by feeding it with a complete dataset, to satisfactory prove its generalization capability. There are, however, many cases scenarios in which it is intriguing to engage in a more complex discussion with respect to how a model can be assessed when data is not found at a single point in space and time. Actually, decentralized and distributed learning techniques have being recently gaining traction in fields where data cannot be shared or moved~\cite{verbraeken2020survey},\cite{AltilioDist}. 

To our knowledge, the field of distributed learning as a whole has still a vast room for improvement since its compelling implementation means have not been fully explored yet, given the rise of ubiquitous data presence. 
As a matter of fact, owing to their peculiar design boundaries, we can consider randomized neural networks as an ideal test bench for exploiting the appealing possibilities of decentralized techniques. In fact, the prevalent advantage of randomized neural networks lies in a much simpler training process while retaining a satisfactory accuracy. 
For this reason, the possibility of extracting information where training is restricted to a portion of the network, enables the study of distributed neural networks implementations.
While shallow networks could have some deficit in large problems, the simplicity of their characteristics can be taken advantage of in the distributed context.
They could be a factor in strengthening the generalization capabilities with respect to local elementary solutions. 
In particular, in an interconnected network of local agents, where each agent is a randomized neural network with access to its own subset of training data,
the randomized weights can be shared amongst the agents. 
Training data available to the agent can be used to train the rest of the local model.
The agent could use its local mode for inference as is but it is expected that it will benefit from getting the additional information about the local models of other agents.
This scenario stems from the following consideration: data is ubiquitous and computational power is scattered all around us.
In other words, it is hard to assume that training data can always be gathered in a single place and it is unlikely to rely on a single centralized agent for handling and analyzing it. 
Therefore, our main focus is to set up a framework in which different agents (i.e., nodes in the network) make use of local training samples to train their own local model, and then sharing some information about their trainable connections (but no training samples) with each other to enhance local predictions. 
Our approach hinges on principles and premises which are also studied in the Federated Learning (FL) framework. 
Namely, the availability of raw data only at local agents (i.e., ``siloed data'') and the impossibility of sharing such raw data are common aspects of both FL and distributed learning. 
In fact, depending on the definition used, there are certain aspects which might differ; in particular, in our work, there is an absence of a ``master'' agent orchestrating the training, which is often present in FL. 
This fundamental distinction makes our work challenging in several aspects (computation, combination) which are specific to the fully distributed framework. 


In this setting, 
the main contribution of this paper is a lossy compression approach based on hyperdimensional computing, which allows us decrease the amount of information being exchanged between the agents.    
As demonstrated by the empirical experiments, the proposed approach achieves a trade-off between the improvement in performance when compared to the local models and communication overheads necessary for information exchange between the agents.



The paper is structured as follows. 
Section~\ref{sec:methods} describes methods used for the proposed approach. 
The experimental setup and materials are presented in Section~\ref{sec:setup}.
Section~\ref{sec:results} is devoted to experimental results on different variations of the proposed approach. 
The results are discussed in Section~\ref{sec:disc}. 
Section~\ref{sec:conc} presents the concluding remarks.

\section{Methods}
\label{sec:methods}

\subsection{Hyperdimensional Computing}

Hyperdimensional computing, also known as Vector Symbolic Architectures (HDC/VSA)~\cite{PlateHolographic2003, RachkovskijStructures2001, MAP, Kanerva2009, Gallant13, FradySDR2020}, is a family of computational frameworks based on random distributed representations,  which are capable of exhibiting the behavior of both a symbolic and a neural nature in a high-dimensional space.
Vectors of high (but fixed) dimensionality (denoted as $D$) are the basis for representing information in HDC/VSA. 
We refer to them as hypervectors. 
The information is distributed across hypervectors components, therefore, hypervectors use distributed representations~\cite{Hinton1986}, which are contrary to the localist representations~\cite{GelderDistributed1999} since any subset of the components can be interpreted. 
It is worth noting that an important property of high-dimensional spaces is that, with an extremely high probability, all random hypervectors are approximately orthogonal to each other.

HDC/VSA also defines a set of operations to manipulate hypervectors. 
In this paper, we use  two key operations: binding and superposition.
The binding operation is used to associate two hypervectors together. The result of binding is another hypervector. 
Here, we will use two implementations of the binding operation. 
First, in the Multiply-Add-Permute framework~\cite{MAP}, the result of binding (denoted as $\textbf{z}$) two hypervectors $\textbf{x}$ and $\textbf{y}$  is calculated as follows: 
$\textbf{z} = \textbf{x}  \odot \textbf{y}$, 
where $\odot$ denotes the binding operation, since it is implemented by  component-wise multiplication.
Another implementation via the circular convolution is presented in Section~\ref{sec:comp}.
An important property of the binding operation is that the resultant hypervector $\textbf{z}$ is approximately orthogonal to the hypervectors being bound.

The superposition operation combines several hypervectors into a single hypervector. 
In contrast to the binding operation, the hypervector resulting from the superposition operation is similar to all component hypervectors, which allows storing information in hypervectors~\cite{Frady17, KleykoPerceptron2020}.
Its simplest realization is a component-wise addition. 
We will use this realization for the compression procedure in Section~\ref{sec:comp}. 
Other realizations of the superposition operation would usually involve the component-wise addition the first step. 
The disadvantage of the component-wise addition is that the vector space becomes unlimited so it is often practical to limit the range of values in the resultant hypervector.
This, for example, can be done with  a clipping function denoted as $f_\kappa ( * )$:
\noindent
\begin{equation}
f_\kappa (x) = 
\begin{cases}
-\kappa & x \leq -\kappa \\
x & -\kappa < x < \kappa, \\
\kappa & x \geq \kappa
\end{cases}
\label{eq:clipping}
\end{equation}
\noindent
where $\kappa$ is a configurable threshold.
The implementation via the clipping function is in particular useful for resource-efficient variants of neural networks such as Self-Organizing Maps~\cite{intSOM} and Echo State Networks~\cite{NepomnyashchiyHardwareESN2020, KleykointESN2017}.
This implementation is also used for a randomized neural network presented in the next section. 

The above operations applied to distributed representations allow using HDC/VSA to produce vector associations that represent, e.g., compositional structures such as sequences~\cite{Kanerva2009},  sets~\cite{KleykoABF2020}, state automata~\cite{YerxaUCBHD_FSA2018, OsipovHD_FSA2017}, hierarchies, or predicate relations~\cite{PlateHolographic2003, KleykoFlyBee2015, RachkovskijStructures2001}. 
Please consult~\cite{KleykoComputingParadigm2021} for a general overview. 
What is more important here is that HDC/VSA can solve a variety of learning tasks with comparable performance to conventional machine learning algorithms~\cite{RahimiBiosignal2019, Kleyko2018,GeClassificationReview2020} or alternatively hypervectors can be used as an input to conventional machine learning algorithms~\cite{RachkovskijClassifiers2007, PSI19, BandaragodaTrajectoryTraffic2019, ShridharEnd2End2020, HyperEmbed}.

In the next subsection, we present a particular example of how HDC/VSA were applied to modify a known  machine learning algorithm.

\begin{figure}[!t]
    \centering
    \includegraphics[width=1.0\columnwidth]{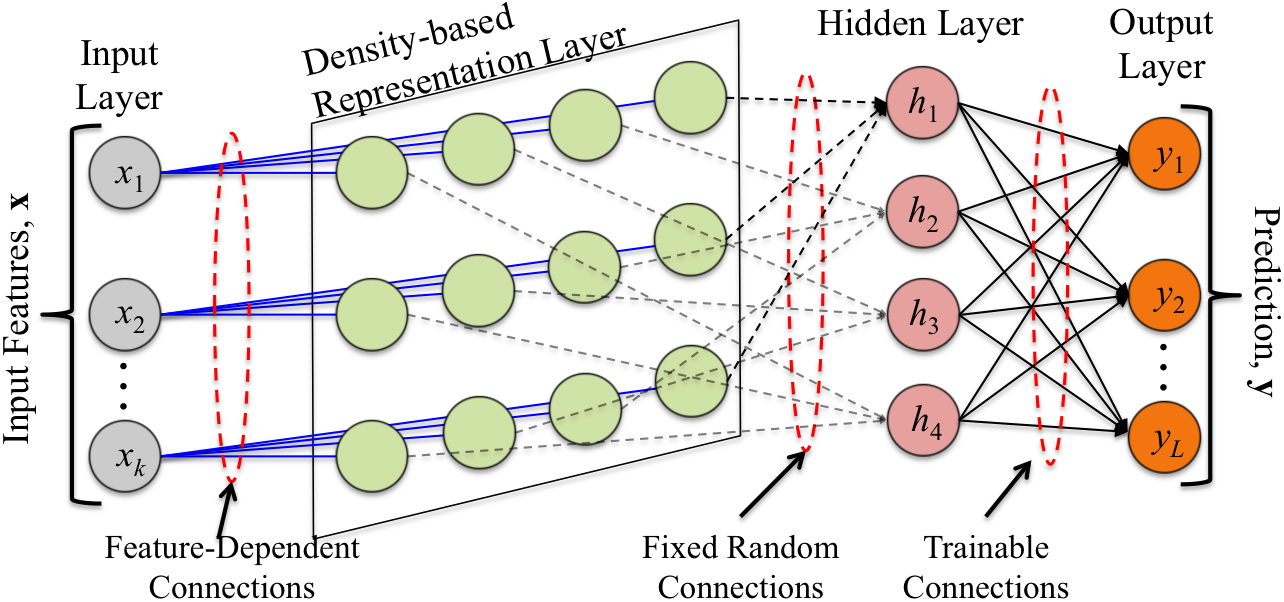}
    \caption{Overview of the RVFL network used in this work. 
    }
    \label{fig:approach}
\end{figure}

\subsection{Random Vector Functional Link networks}

As a model for randomized neural networks, here we use Random Vector Functional Link (RVFL) networks~\cite{RVFLorig} also known as Extreme Learning Machines~\cite{ELM06}.
In these feedforward neural networks, connections between the input and hidden layers are set at random and fixed during network's lifetime~\cite{Scardapane2017}.
In the experiments, we used a recent modification of RVFL networks from~\cite{intRVFL2020}, which is based on the ideas from HDC/VSA to compute the activation values of the hidden layer from input features.
Fig.~\ref{fig:approach} illustrates the network's architecture. 

The network begins by quantizing the values $\textbf{x}_i$ of individual input features from the input sample $\textbf{x} \in [K \times 1]$.
Each feature is then represented in the form of $D$-dimensional bipolar hypervector using the thermometer code~\cite{Scalarencoding}.
These hypervectors are collected in matrix $\mathbf{F} \in [D \times K]$.
This step is called density-base representation layer in Fig.~\ref{fig:approach}.
The thermometer codes are going to be bound with random (but fixed) bipolar hypervectors assigned to each feature at the initialization step of the RVFL network.
These $K$ $D$-dimensional hypervectors are stored in a matrix $\mathbf{W}^{\mathrm{in}} \in [D \times K]$.
In Fig.~\ref{fig:approach}, $\mathbf{W}^{\mathrm{in}}$ is indicated as fixed random connections between the density-base representation and hidden layers. 
For the $j$th feature the result of the binding operation is simply  $\mathbf{W}^{\mathrm{in}}_j  \odot \mathbf{F}_j$. 
This results gets fed into the hidden layer.
The full activation of the hidden layer (denoted as $\mathbf{h} \in [D \times 1]$) is computed as: 
\noindent
\begin{equation}
\mathbf{h}  = f_\kappa \left( \sum_{j=1}^{K} \mathbf{W}^{\mathrm{in}}_j  \odot \mathbf{F}_j  \right).
\label{eq:hidden}
\end{equation}
\noindent
Note that the clipping function here acts as an activation function and introduces the nonlinearity.
Since the connections prior to the hidden layer are fixed, RVFL networks 
need to train only the classifier part of the network (trainable connections in Fig.~\ref{fig:approach}) located between the hidden and output layers. 
The next subsection describes this step.

\subsection{Classifiers}
\label{sec:class}

In this paper, we study two ways of forming the classifier (denoted as $\mathbf{W}^{\mathrm{out}}$) of the representations produced by the hidden layer.
The first way is standard in RVFL networks~\cite{RVFLorig}: it uses the result of the regularized least squares applied to hidden layer activations. 
The second way is common in HDC/VSA and relies on class centroids, which are formed by simply superimposing all hidden layer activations for the corresponding class. 
In our opinion, it is interesting to analyse the generalization capabilities of these two different approaches because they reach the solution to the classification problem by carrying out radically different operations. In particular, the centroid classification exploits the VSA high dimensionality to obtain $\mathbf{W}^{\mathrm{out}}$ using a simpler, more elementary method. This might be especially desirable in resource-constrained devices.

Once $\mathbf{W}^{\mathrm{out}}$ is formed, to classify a given input sample $\mathbf{x}$, the RVFL network first produces the activations of the hidden layer $\mathbf{h}$.
Then it computes the possible classification labels $\hat{\mathbf{y}}$, corresponding to the activations of the output layer, as the dot product between $\hat{\mathbf{y}}$ and $\mathbf{W}^{\mathrm{out}}$:  $\hat{\mathbf{y}} = \mathbf{W}^{\mathrm{out}} \mathbf{h}$. 
The predicted class corresponds to a component in $\hat{\mathbf{y}}$ whose activation is the largest; this mechanism is known as winner-takes-all.
We now describe the two classifiers studied herein, which will be at the core of the experiments reported in Sec. \ref{sec:setup} and \ref{sec:results}.

\subsubsection{Regularized least squares classifier}

The output layer of an RVFL network corresponds to a classifier.
The standard way to formulate the problem of obtaining the optimal values in $\mathbf{W}^{\mathrm{out}}$ is by minimizing the mean squared error between the ground truth and the output of the RVFL network. 
When solving classification problems, the ground truth is represented as one-hot vectors of the corresponding class labels (denoted as $\mathbf{y} \in [L \times 1]$), where $L$ denotes the number of classes in the problem. 
Activations of the hidden layer ($\mathbf{h}$) for all $M$ training samples are collected as rows in an activation matrix $\mathbf{H} \in [M \times D]$. 
While the corresponding ohe-hot encodings $\mathbf{y}$ are collected in another matrix $\mathbf{Y} \in [M \times L]$. 

Following this reasoning, the regularized least squares (RLS) classifier can be obtained in one analytic step using $\mathbf{H}$  and $\mathbf{Y}$ as: 
\begin{equation}
\mathbf{W}^{\mathrm{out}} = (\mathbf{H}^{\top} \mathbf{H} + \lambda \mathbf{I})^{-1} \mathbf{H}^{\top} \mathbf{Y},
\label{eq:rls}
\end{equation}
\noindent 
where $\lambda$ is a regularization parameter; $\mathbf{I} \in [D \times D]$ is the identity matrix. 
$D \times D$ matrix inverse dominates the computational complexity of (\ref{eq:rls}), which might be rather intense especially when $D$ is getting large.

\subsubsection{Centroids classifier}

In HDC/VSA, the superposition operation is one of the key operations, so it is widely used for classifiers using centroids.
In this case, the classifier $\mathbf{W}^{\mathrm{out}}$ consists of individual centroids where $\mathbf{W}^{\mathrm{out}}_i$ denotes the centroid for class $i$. 
The main idea with centroids is that they might provide high between-class variability (i.e., centroids for different classes are dissimilar to each other) while they would also have low within-class variability~\cite{Daugman2003}.
In other words, it is expected that hidden layer activations of class $i$ are going to be very similar to their class centroid  $\mathbf{W}^{\mathrm{out}}_i$. 
A centroid of a class is computed simply from the hidden layer activations of the training samples, which belong to that class, as follows:

\begin{equation}
\mathbf{W}^{\mathrm{out}}_i = \frac{\sum_{\mathbf{h}^{(t)} \in i} \mathbf{h}^{(t)}}{\left\Vert \sum_{\mathbf{h}^{(t)} \in i} \mathbf{h}^{(t)}  \right\Vert_2},
\label{eq:cent}
\end{equation}
\noindent 
where $\mathbf{h}^{(t)} \in i$ denotes that $t$th training sample is used to compute the centroid for class $i$ only if this sample belongs to class $i$. 
The normalization is used to account for the fact that different classes might be represented by different number of training samples.
In~\cite{DiaoGLVQHD2021}, the centroids were used as an initial step for a more general classifier known as Generalized Learning Vector Quantization~\cite{Nova2013} but here we limit ourselves to centroids only.


\subsection{Distributed classification}
\label{sec:dist}

In Section~\ref{sec:intro}, we already touched the reasons why a distributed approach is worth exploring. In this work, we limit our scope to the case where a set of agents is scattered in the space, forming a network.
Each agent has its own computation and communication capabilities.
Every agent gets its own local subset from a full dataset spread over the network. 
In our case, we consider these subsets being sampled independently without replacement from the full dataset\footnote{With this regard, other mechanisms can be used. 
We have chosen this as being the most straightforward one, giving us the ability to distill the consequence of the use of centroids classifier in the distributed scenario without introducing variations that would impede the direct comparison with the RLS classifier}. Moreover, the agents cannot share their training samples as it would hinder the pure distributed definition of the problem.
They, however, are allowed to share information about their classifiers with each other.

Formalizing, we consider the training data $\mathcal{T}$ being distributed over a network of $N$ interconnected agents. 
The network of agents can be modeled as a directed graph $\mathcal{G}( \mathcal{V}, \mathcal{E} )$, where $\mathcal{V} = \{1, \, \ldots, \, N \}$ is the set of the agents, and $\mathcal{E}$ is the set of the edges. The connectivity of the graph is fixed and known a priori and can be described as a $N \times N$ adjacency matrix denoted as
$\mathbf{\Omega}$; where $\mathbf{\Omega}_{p,q} \neq 0$ if and only if agents $p$ and $q$ are connected. 
While there are several strategies to choose the weights of the connections, in the following, for the sake of simplicity, we will consider the undirected, fully connected graph ($\mathbf{\Omega}_{p,q} = 1, \, \forall p,q \, \in \mathcal{V}$).
Also, for the sake of working in a fully-distributed framework, we impose three additional constraints in the design of our approach:
\noindent
\begin{itemize}
\item no agent is allowed to coordinate the training process;
\item agents are not allowed sharing data samples;
\item communication is allowed only between connected agents.
\end{itemize}
\noindent
These principles do not restrict the proficiency of our proposal in its suitability to be employed in different distributed applications.

Next, we detail the means by which the distributed training procedure is carried out. 
First, each agent $p$ obtains its own classifier
$\mathbf{W}^{\mathrm{out}(p)}$ using agent's training samples. 
Second, each agent $p$ has a set of neighbor agents $\mathcal{S}$ that are all the agents $s$ for which $\mathbf{\Omega}_{p,s}\neq0$. 
Given the limitations placed for the purely distributed case, the crucial step is now for agent $p$ to collect the information about the classifiers of its neighbor agents $\mathbf{W}^{\mathrm{out}(s|s\in\mathcal{S})}$.
Once these classifiers are collected, the agent can combine them together into an aggregated classifier, $\mathbf{W}^{\mathrm{dist}(p)}$. 
The aggregation is done by simply summing up all the  classifiers (including agent's own one) received from the neighbors:
\noindent
\begin{equation}
   \mathbf{W}^{\mathrm{dist}(p)}=  \sum_{s \in \mathcal{S}} \mathbf{W}^{\mathrm{out}(s)}.
\end{equation}
\noindent
We do not apply any iterative approach, avoiding to analyse the convergence by using something similar to a one shot average.
In the proposed approach we consider two different ways of exchanging $\mathbf{W}^{\mathrm{out}(p)}$ between the agent. 
In the most straightforward case, they are shared as is. 
A more subtle case is when $\mathbf{W}^{\mathrm{out}(p)}$ is sent in the compressed form to save communication resources. 
The next subsection elaborates on the compression procedure.

\subsection{Compression of the classifier}
\label{sec:comp}

\begin{figure}[!t]
    \centering
    \includegraphics[width=0.7\columnwidth]{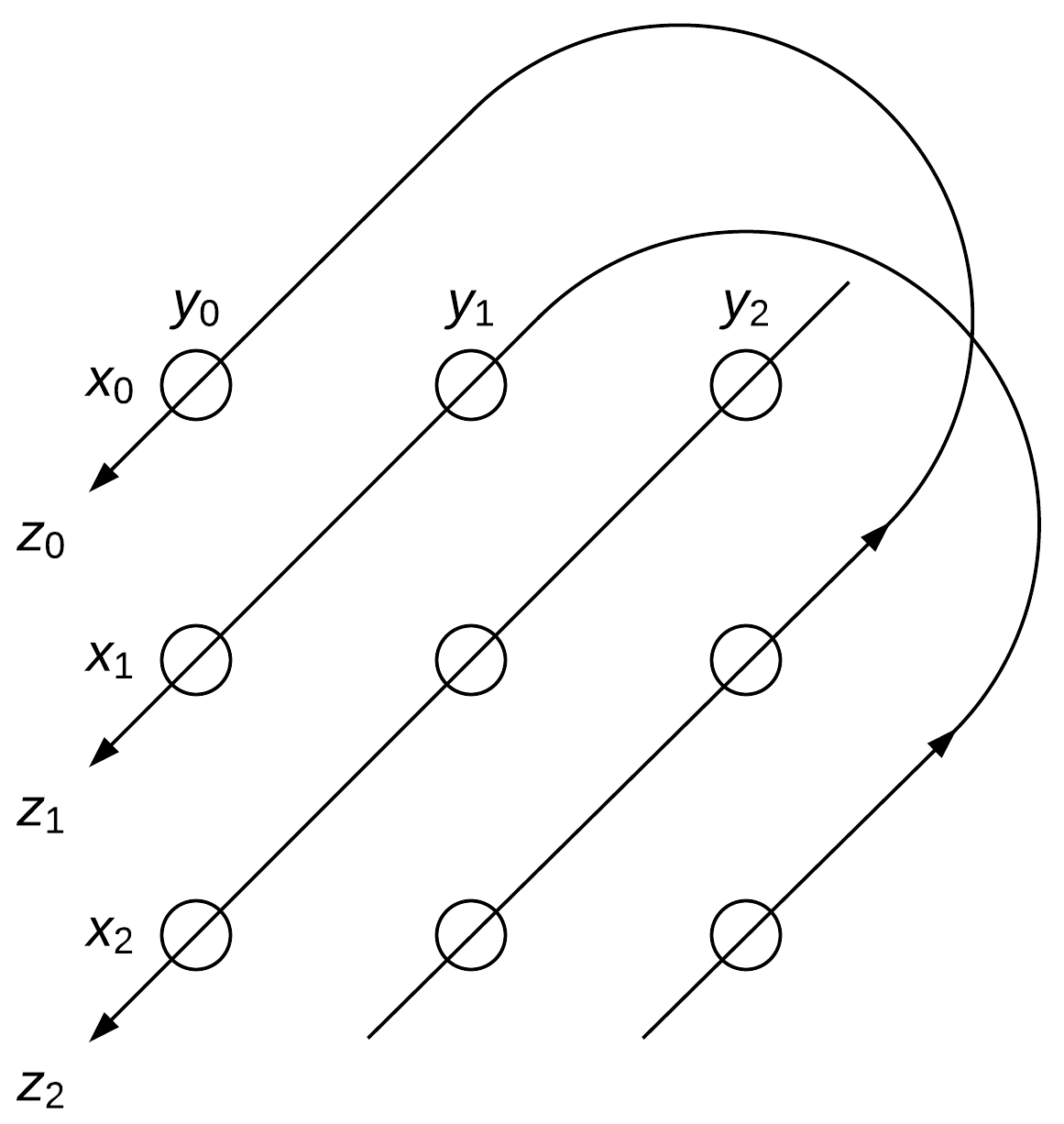}
    \caption{An example of the circular convolution operation. 
    }
    \label{fig:circ:conv}
\end{figure}

As discussed in the previous section, agents need to exchange information about locally computed versions of $\mathbf{W}^{\mathrm{out}(p)}$.
Since communicating information can be a costly operation we are interested in minimizing it. 
HDC/VSA can be seamlessly used to compress $\mathbf{W}^{\mathrm{out}}$ into a single $D$ dimensional vectors. 
To do so we use HRR framework~\cite{PlateHolographic2003}.
The first step is to form $L$ hypervectors (do not confuse with $\mathbf{W}^{\mathrm{in}}$) corresponding to key-value pairs where in each pair a value is a classifier for class $i$, $\mathbf{W}^{\mathrm{out}}_i$ while a key is a random hypervector corresponding to class $i$ (denoted as $\mathbf{K}_i$).
The key-value pair hypervector is formed via the binding operation, which in the HRR framework is implemented via the circular convolution. 
The circular convolution (denoted as $\circledast$) can be seen as a compressed version of the outer product of vectors being bound. 
Fig.~\ref{fig:circ:conv} shows an example for three dimensions when performing the binding: 
\noindent
\begin{equation*}
\mathbf{z}=\mathbf{x} \circledast \mathbf{y}.
\end{equation*}
\noindent
The individual components of the result in $\mathbf{z}$ are calculated as:
\noindent
\begin{equation*}
\mathbf{z}_0=\mathbf{x}_0\mathbf{y}_0+\mathbf{x}_2\mathbf{y}_1+\mathbf{x}_1\mathbf{y}_2;
\end{equation*}
\noindent
\noindent
\begin{equation*}
\mathbf{z}_1=\mathbf{x}_1\mathbf{y}_0+\mathbf{x}_0\mathbf{y}_1+\mathbf{x}_2\mathbf{y}_2;
\end{equation*}
\noindent
\noindent
\begin{equation*}
\mathbf{z}_2=\mathbf{x}_2\mathbf{y}_0+\mathbf{x}_1\mathbf{y}_1+\mathbf{x}_0\mathbf{y}_2.
\end{equation*}
\noindent
In general case, the value of the $j$th component is calculated as: 
\noindent
\begin{equation*}
\mathbf{z}_j= \sum_{k=0}^{D-1}  \mathbf{y}_k \mathbf{x}_{j-k \mod D}
\end{equation*}
\noindent
Thus, in our scenario, we form $L$ (which has been already defined as the number of classes) bound hypervectors: $\mathbf{K}_i \circledast \mathbf{W}^{\mathrm{out}}_i$.
These hypervectors are used to form the compressed version of $\mathbf{W}^{\mathrm{out}}$ (denoted as $\mathbf{w}$), which contains their superposition: 
\noindent
\begin{equation*}
\mathbf{w}= \sum_{i=1}^L \mathbf{K}_i \circledast \mathbf{W}^{\mathrm{out}}_i.
\end{equation*}
\noindent
Hypervector $\mathbf{w}$ is the compression version of the classifier, which is going to be sent to other agents when exchanging the information in order to enhance the performance of local classifiers.
It is worth noting that it is assumed that each agent will generate its own random matrix $\mathbf{K}$ storing key hypervectors but all agents will be able to regenerate $\mathbf{K}$ on their own. 
This is not an unrealistic assumption since $\mathbf{K}$ can be generated using, e.g., agent's ID as a seed to initialize a pseudorandom number generator. 

When an agent receives $\mathbf{w}$ from some other agent $q$ it has perform the decompression procedure to reconstruct the approximate version of $\mathbf{W}^{\mathrm{out}}$. 
The decompression is done for each $\mathbf{W}^{\mathrm{out}}_i$ using the inverse of the corresponding key hypervector\footnote{Please refer to~\cite{PlateHolographic2003} for the details of constructing the inverse of $\mathbf{K}_i$.} of the agent $q$ as follows:
\noindent
\begin{equation*}
\hat{\mathbf{W}}^{\mathrm{out}}_i \approx \mathbf{w}  \circledast \mathbf{K}_i^{-1}.
\end{equation*}
\noindent
The reconstructed classifier $\hat{\mathbf{W}}^{\mathrm{out}}$ is not going to be the exact replica of the original $\mathbf{W}^{\mathrm{out}}$ since the compression with the superposition operation is lossy, as other key-value pairs act as a crosstalk noise during the reconstruction process. 
Nevertheless, our hypothesis is that when combining $\hat{\mathbf{W}}^{\mathrm{out}}$ from different agents, their crosstalk noise will average out without having a large effect on the classification results. 
Thus, this approach should act as a trade-off between the advantages of exchanging the information amongst agents and the communication overheads associated with this exchange.

\section{Experimental setup}
\label{sec:setup}

\subsection{Materials}
\label{sec:data}

For assessing the performance of the proposed approach, we carried out experiments on a collection of $121$ benchmark classification datasets from the UCI Machine Learning Repository~\cite{Dua2019}.
The collection was prepared as a part of the seminal work~\cite{Delgado2014}. 
Here, we followed the same experimental protocol as in  ~\cite{Delgado2014}.
The only addition we introduced to the preprocessing of the datasets was the normalization of input features to the range $[0, 1]$ prior to providing input samples to a network. 
There was no additional preprocessing other than that.
In the experiments involving only the centralized version, we used the full collection. 
In experiments involving distribution of the datasets between agents, we had the need to discriminate the datasets based on their size.
This was done by setting up a threshold to ensure there were at least some data points in each and every local subset. 
Therefore, we used only datasets with more than $1,000$ samples in the training part. 
There are $42$ such datasets in the collection. 


\subsection{Setup}

In order to obtain a good grasp on what to expect from the proposed approach for distributed classification, we experimented with three different classification models introduced below.

\paragraph{Centralized version} this model simulates the situation where all training samples are gathered in a central location. 
The dataset is, thus, analyzed as a whole, and the performance is evaluated on a single RVFL network. 
It should be noted that this version is an extreme case benchmark where all the information is available to a single agent. 
Therefore, this model can be considered as a kind of ``upper-bound'' benchmark, in the sense that the other two versions below are expected to perform worse or on a par with it, given their limited access to the data. 
As explained in Section~\ref{sec:intro}, there are practical scenarios where the centralized version might be infeasible.

\paragraph{Local version} the most elementary implementation of the proposed approach that we analyzed stems from the scattering of the computational power in the network of interconnected agents. 
In this case, each agent in the network is considered to have access only to local subset of the dataset, without the possibility of sharing any information with its neighbors. 
For evaluation purposes, in this version each subset of dataset is taken randomly without replacement from the full dataset, mimicking the real-world decentralized case.

\paragraph{Distributed version} this case represents the realization of the proposed approach to distributed classification, where the information flow among agents is restricted to the means already detailed in Sec. \ref{sec:dist}.
We allow agents sharing their locally computed classifiers $\mathbf{W}^{\mathrm{out}}$ either as is or after a compression procedure (detailed in Section~\ref{sec:comp}).    
Also, in the case studied here, all agents can communicate with each other (i.e., the network is fully connected). 
This way enables isolating the effect of learning on decentralized dataset without taking into account additional consequences stemming from different protocols for sharing the information. 

\subsection{Hyperparameters}

The search of the ($D$, $\lambda$, and $\kappa$) was done according to~\cite{Delgado2014} with the grid search for each dataset using the RLS classifier and the centralized version. 
$D$ varied in the range $[50, 1500]$ with step $50$;  $\lambda$ varied in the range $2^{[-10, 5]}$ with step $1$; and $\kappa$ varied between $\{1,3,7,15\}$.
The chosen values were used for both classifiers and all the versions reported in the next section. 

It is worth noting that for all considered models, it was assumed that the agents share the same values of $\mathbf{W}^{\mathrm{in}}$, which were chosen equiprobably from $\{-1,+1\}$. 
Practically, it is easy to ensure the same $\mathbf{W}^{\mathrm{in}}$ by letting all agents to share the same seed for their pseudorandom number generators.
In order to avoid the influence of a particular random selection of $\mathbf{W}^{\mathrm{in}}$ on model's performance, all results reported below were averaged for $10$ random initializations of $\mathbf{W}^{\mathrm{in}}$.


\begin{figure}[t]
    \centering
    \includegraphics[width=1.0\columnwidth]{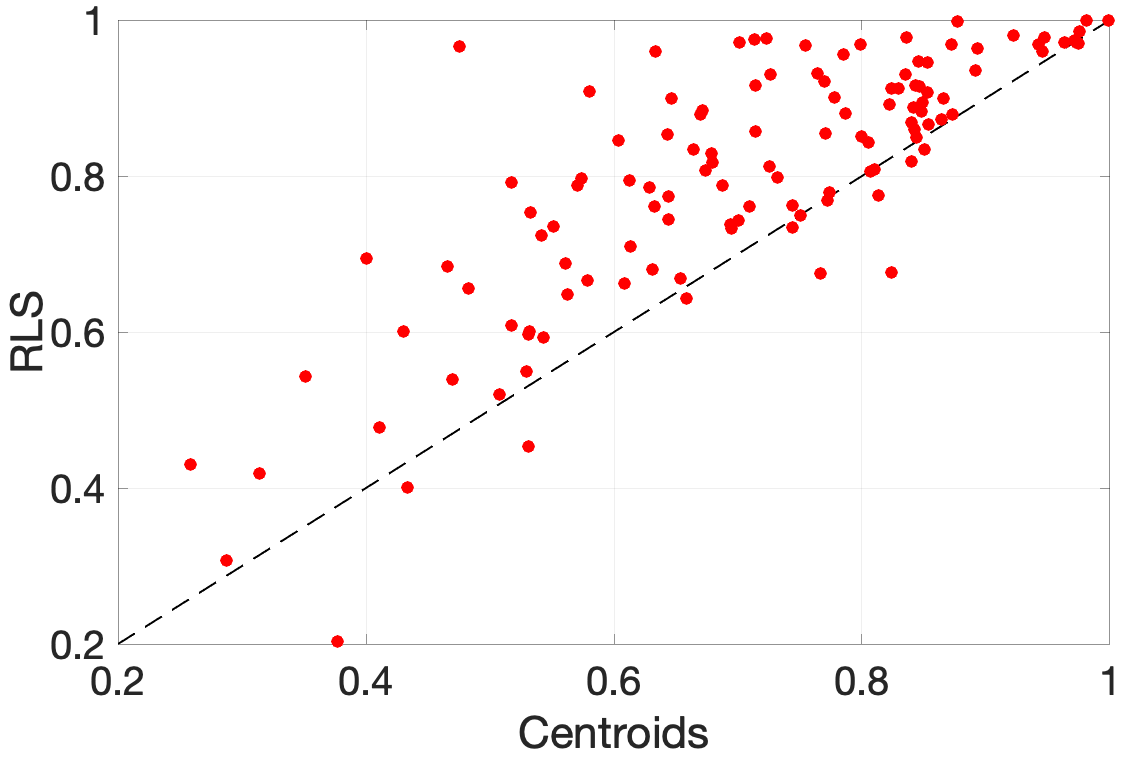}
    \caption{The cross-validation accuracy of RLS classifier (mean $0.80$) against the centroids classifier (mean $0.70$) on all $121$ datasets. Each point corresponds to a dataset.
    The results were averaged over $10$ simulation runs. 
    }
    \label{fig:exp:rls:cent}
\end{figure}

\begin{figure*}[t]
    \centering
    \includegraphics[width=2.0\columnwidth]{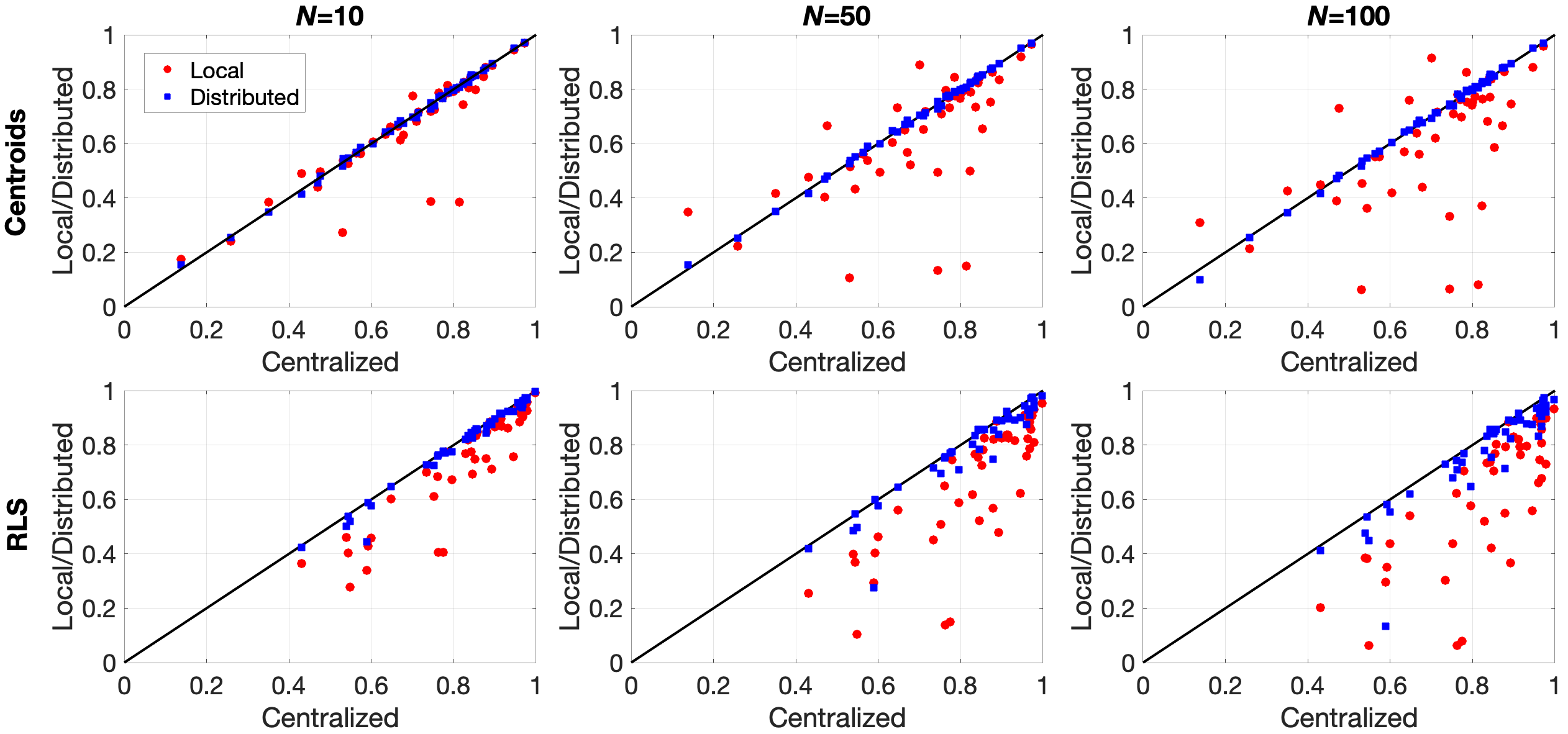}
    \caption{The cross-validation accuracy of the local and distributed version vs. the centralized version for three different number of agents ($N \in \{10, 50, 100\}$). There was no compression for the distributed version.
    Upper panels: centroids classifier;
    Lower panels: RLS classifier.
    Each point corresponds to a dataset.
    The results were averaged over $10$ simulation runs.
    }
    \label{fig:exp:central:distr}
\end{figure*}

\section{Experiments}
\label{sec:results}

\subsection{RLS vs. centroids in the centralized setup}

In this subsection, our goal is to compare the two classifiers considered in this study: RLS and centroids. 
To do so, we use the most standard setup -- the centralized version.  
Since data is not distributed between agents, in this experiments we were able to use the full collection.

Fig.~\ref{fig:exp:rls:cent} presents the results in the form of the cross-validated accuracies obtained by each of the classifiers. 
The Pearson correlation coefficient between the results obtained from the classifiers was $0.80$, which indicates high dependence between the results. 
We could clearly see, however, that the RLS would usually outperform the predictions obtained from the centroids.
In this regard, the average accuracy for the RLS was $0.80$ while that for the centroids was only $0.70$.
At the same time, it is worth pointing out that there were numerous datasets where the centroids performed on a par with the RLS. 
For example, there were $32$ datasets where the accuracy of the centroids was worse to less than $0.02$ of the RLS accuracy. 
In other words, for approximately a quarter of the datasets, the centroids classifier was a viable option.
This indicates that while being very simplistic, the centroids classifier could still be a useful approach.

\subsection{Local and distributed vs. centralized versions: no compression}

In this subsection, our goal is to assess the difference between the local and distributed versions. We used the centralized version of the corresponding classifier as a baseline.
No compression was used in the experiments reported herein.  
To get a grasp on the performance of the distributed paradigm, we considered three different experiments, varying only the number of agents in the network $N$, in the set $\{10, 50, 100\}$. 
For every experiment, we used local independent subsets for both training and test, and then computed the average accuracy over the whole network of agents. 
In Fig.~\ref{fig:exp:central:distr}, the results are reported for the three values of $N$ (columns in Fig.~\ref{fig:exp:central:distr}) and two different classifiers (rows in Fig.~\ref{fig:exp:central:distr}). 
The results are based on $42$ datasets selected from the collection as described in Section~\ref{sec:data}.

\begin{table}[tb]
\renewcommand{\arraystretch}{1.0}
\caption{Average accuracies for different versions and classifiers.
}
\label{tab:accuracices}
    \begin{center}
    \begin{tabular}{|c | c | c | c | c | c |}
     \cline{3-6} 
    \multicolumn{2}{c|}{ } & $N=1$ & $N=10$ & $N=50$ & $N=100$  \\ \hline
    \multirow{2}{*}{Cent.} & Local &  $0.70$ & $0.67$ &  $0.63$ &  $0.60$ \\ \cline{2-6} 
    & Distr &  $0.70$ & $0.70$ &  $0.70$ &  $0.70$ \\ \hline 
    \multirow{2}{*}{RLS} & Local &  $0.83$ & $0.74$ &  $0.66$ &  $0.62$ \\ \cline{2-6} & Distr &  $0.83$ & $0.82$ &  $0.80$ &  $0.78$ \\ \hline      
    \end{tabular}
    \end{center}
\end{table}

The results allow us making several interesting observations. 
The main one is that the distributed version clearly performed better than the local one, bringing the performance very close to the centralized version. This is important since it justifies that the exchange of information between the agents had its positive impact for the classification performance. 
Getting into more specific observations, we see that, for both classifiers, local versions are getting noticeably worse with increased $N$. 
Table~\ref{tab:accuracices} reports the average accuracies. 
Note that $N=1$ corresponds to the centralized version and local and distributed versions are equivalent to each other.
This tendency is not unexpected since the data was split between the agents without replacement, which means that the amount of training samples provided to a single agent was decreasing with increased $N$.
When it comes to the distributed classification, the results for the centroids classifier always matched the centralized version. 
This result is also expected since in our setup after exchanging their local centroids, each agent should obtain the exact replica of the centralized version classifier. 
The situation is more subtle with the RLS classifier.
It is clear that for any number of agents the distributed version is better than the local one (see Table~\ref{tab:accuracices}).
Interestingly, the relative improvement of the distributed version over the corresponding local version was increasing with $N$: $10.1$\%, $21.1$\%, and $25.8$\%, respectively. 
At the same time, we see that the results were decreasing with increased $N$.
That is because aggregating together local RLS classifiers is not equivalent to computing a single classifier from the full dataset (the average accuracy of the centralized version was $0.83$) and, therefore, we do not see exactly the same performance as in the case of the centroids classifier.

\subsection{Distributed version with compression}

The fact that the distributed version without compression compares favorably to the local version makes it interesting to evaluate the performance of the distributed version with compression and assess it in relation to the results of the previous experiment. 

\begin{figure*}[t]
    \centering
    \includegraphics[width=2.0\columnwidth]{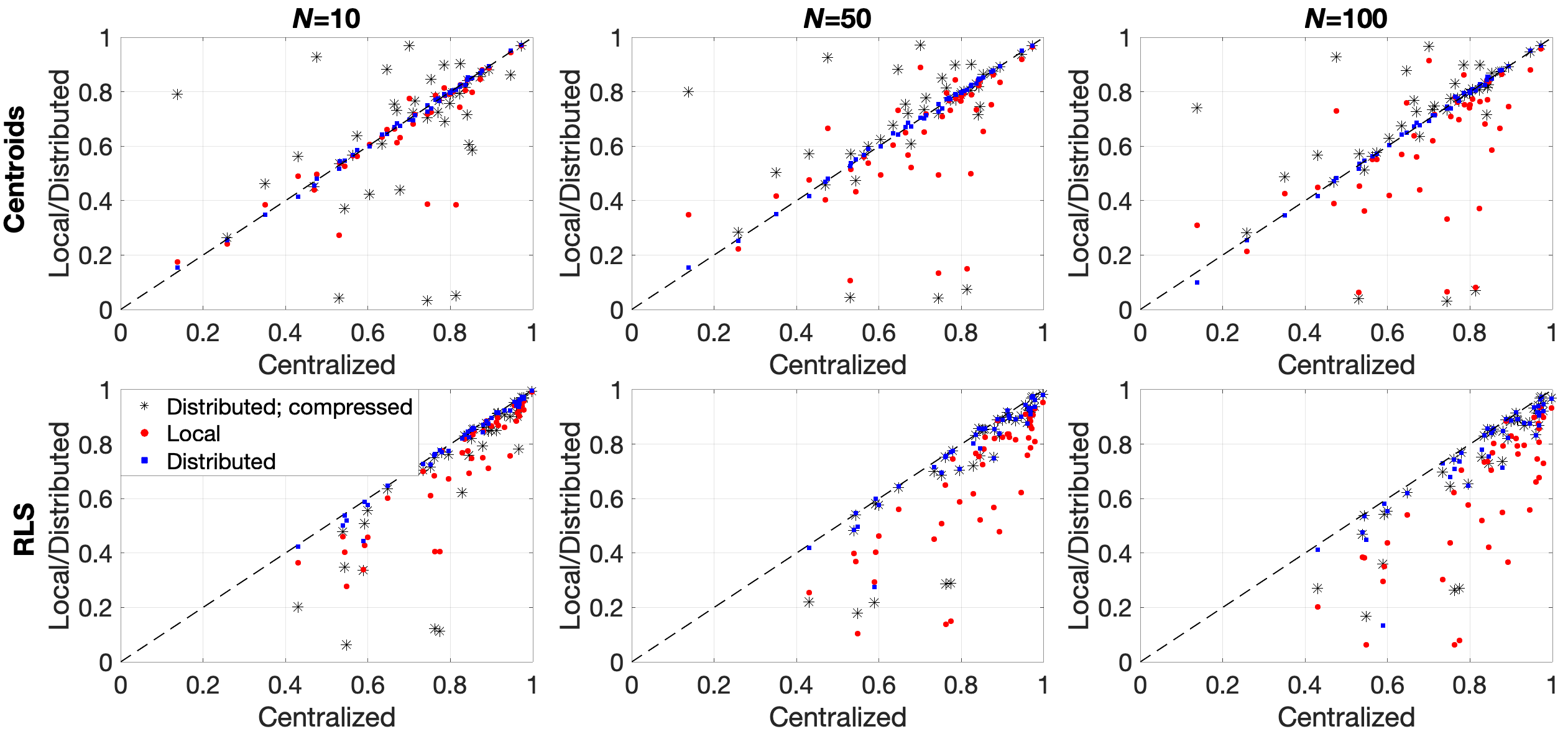}
    \caption{The cross-validation accuracy of the local and distributed version vs. the centralized version for three different number of agents ($N \in \{10, 50, 100\}$). 
    The results for the distributed version were obtained using the compression procedure from Section~\ref{sec:comp}.
    Upper panels: centroids classifier;
    Lower panels: RLS classifier.
    Each point corresponds to a dataset.
    The results were averaged over $10$ simulation runs.
    }
    \label{fig:exp:central:distr:comp}
\end{figure*}

Fig.~\ref{fig:exp:central:distr:comp} reports the results using the same datasets and format of presentation as in Fig.~\ref{fig:exp:central:distr}.
Circle and square markers correspond to the results from the previous experiment for local and distributed (without compression) versions, respectively.
Asterisk markers correspond to the results for the distributed version with compression. 

The centroids classifier provided some unexpected results since for some datasets the accuracy was higher than that of the centralized version. 
We conjecture that the cause of this effect should be the fact that for some of the datasets, the noise introduced by the compression procedure acted as some kind of regularization.  
Nevertheless, there were also datasets for which the distributed version with compression performed worse than the local one. 
However, the average accuracy for $N=10$ was the same as for the local version ($0.67$). 
For larger values of $N$ the average accuracy was higher than for $N=10$ ($0.70$ for both $N=50$ and $N=100$), which demonstrated the tendency opposite to the local version where the average accuracy decreased for larger $N$. 
Moreover, the average accuracy of the distributed version with compression for $N=50$ and $N=100$ was the same (i.e., $0.70$; Pearson correlation coefficient: $0.93$) as for the distributed without compression and centralized versions.  

Similar to the centroids classifier, the average results for the RLS classifier for $N=10$ in the case of compression were the same as for the local version ($0.74$; Pearson correlation coefficient: $0.93$). 
The results have, however, improved for larger values of $N$ to $0.76$ and $0.75$, respectively.
This results have the same trend as in the case of the distributed version without compression: the relative improvement over the local version was increasing with $N$: $0.0$\%, $15.2$\%, and $20.1$\%, respectively. 
Moreover, the difference in performance between the distributed version with and without compression was shrinking with $N$ as the version without compression was better by $10.1$\%, $5.3$\%, and $4.0$\%, respectively.

\section{Discussion}
\label{sec:disc}

\subsection{Related work}

In this work, we have considered two classifiers.
The RLS classifier is a standard choice for the RVFL networks.
The centroids classifier is less common because, as we have seen in the experimental results, its average accuracy was lower than that of the RLS classifier.
At the same time, there were quite a few datasets, where the centroids classifier demonstrated comparable accuracy. 
When taking into account its simplicity, this explains why this classifier is often a common choice in HDC/VSA literature~\cite{Rahimi2016Lang, Najafabadi2016, Rahimi2016EMG, Rasanen2014, Yilmaz2015}.
Nevertheless, there is understanding that the centroids are not always the best choice of the classifier, which motivates to refine them using, e.g., the perceptron learning rule~\cite{Imani2017, Imani2019AdaptHD, Yeseong2018}.
A more general approach to classification, which is based on centroids, is Learning Vector Quantization~\cite{Nova2013}. 
It has recently been introduced in the context of HDC/VSA~\cite{DiaoGLVQHD2021} and it will be important to explore how the proposed approach will deal with the classifiers obtained via Learning Vector Quantization.

The compression procedure is similar in its spirit to the recent idea of using the binding and superposition operations of HDC/VSA to represent parameters of many deep neural networks in a single hypervector~\cite{CheungSuperposition2019}. 
The difference in the presented procedure is that the classifier was reconstructed back from a single hypervector, which was not the case in~\cite{CheungSuperposition2019}.
The attempts to apply HDC/VSA in the communication domain~\cite{JakimovskiCollective2012, KleykoMACOM2012, KimHDM2018, SimpkinHDWorkflow2019} are also related but there the goal would usually be to extract the data back from the hypervector without any losses, which is not the case in our scenario.

Regarding the distributed classification domain, there are some works worth mentioning, which are related to the proposed approach.
First of all, let us discuss the works pertaining to distributed classification. 
In this field, most of the authors presented solution that can be ascribed to the broad field of wireless sensor networks. A good review of such applicative methods can be found in \cite{park2020communication}. 

In more specific terms, regarding purely distributed classification (i.e., considering only algorithms that do not imply a master/slave interaction nor sharing of data of any kind), there are several works proposing a detailed analysis of the communication and convergence of the global problem.
For instance, in \cite{zhai2020distributed} a distributed classification is studied, based on the Broad Learning System \cite{chen2017broad}, which is quite complex in its computation, since it involves optimizing the Broad Learning System via Alternative Direction Method of Multipliers (ADMM).  
In \cite{zhang2018distributed} a decentralized classification solution is proposed, similar in is premises to ours, but the focus of the work is on parallel classification performance. 
Another work that focuses on a similar distributed approach, levering a divide-and-conquer logic, is \cite{lin2017distributed}. In effect, such works studying the parallelization of a task on multiple agents in a network are more focused on analysing the scalability and convergence of the problem, veering from the actual distribution.
Still regarding the diffusion of information through the network, in \cite{montero2019distributed} the probability distribution is the focus of this analysis, getting to the point of studying the complexity of the combination of unbalanced agents. 
While interesting in the general distributed framework, we believe that the simplicity of the network computation must be at the core of the distributed learning study.
Similarly, works like \cite{lewis2017cooperative}, in which the transfer of information among agents is carried out via sharing the gradients through backpropagation, are related to the present study, but do not offer a worth comparison in terms of intelligibility and effortlessness.

Moving to the specific distributed randomized network framework, there are some works in which this very scheme is studied. 
For instance, \cite{scardapane2015distributed} was one of the first works in which authors introduced a distributed RVFL scheme, but, differently from us, they derived it for sequential data. 
In \cite{scardapane2015learning}, a very similar technique has been studied, focusing on a single problem distributed in several agents, optimizing it via ADMM. 
As it is evident from this discussion, the ADMM is a well-known solution for distributed problems, and it is used also for more complex distributed network optimization \cite{fierimonte2016distributed}. By contrast, we want to examine not the optimization of the single problem, but how well a single agent can approximate the global result by only sharing some insights on the classification. 
Finally, it is worth mentioning a good framework studying a generalized, sparse, time varying implementation of parallel and distributed learning with neural networks \cite{scardapane2017framework}, to give to the reader a glimpse of a broader perspective on the problem.

In the context of HDC/VSA, there are a couple of studies~\cite{KleykoIndustrial2018, ImaniHDColLearn2019}, which dealt with the case where a training dataset is distributed across the network but, in contrast to this work, both studies assumed some centralization in their approaches.

\subsection{Extension of the current work}
We have made several assumptions when performing the experiments in this study. 
While the current setup is very useful for proof of concept, it is important to extend it in the future work to be able to make stronger claims about the propose approach. 
Below, we indicate the directions for future work.

\paragraph{Strategies for splitting data between agents} In this study, we only considered the case when data was split between agents randomly without replacement. 
While it allowed us abstracting from the questions of exploring strategies for splitting data, this decision had an effect on the results since the average accuracy was decreasing with increased number of agents due to the shrinking number of training sample per agent.
In the future work, we need to consider other scenarios such as  sub-sampling the full dataset with replacement or allowing partial sharing of samples in the agents.

\paragraph{Comparison with standard compression methods} Here, we proposed to use of HDC/VSA operations as a way to compress agent's classifier before sharing it with its neighbors. 
We have compared the results with the version without compression to make a proof of concept of the idea but in the future work we will make a comparison with the standard compression methods in order to see how HDC/VSA-based approach position itself.

\paragraph{Topology of the connectivity graph} In this work, we have assumed fully connected graph as a topology between the agents. 
It is a simple and intuitive topology to work and experiment with but it is not very practical as it is often the case that all-to-all connectivity is not available.  
While reporting results for other topologies falls outside of the scope of this paper, investigating alternative choices of connectivity graph that better emulate real-world scenarios by restricting number of neighbors will be an important direction for future work.
It will also be important to study how the aggregated classifiers converge with iterations of information exchange.

\paragraph{Advanced classifiers}
In this paper, we used only two rather simple classifiers. 
Future work would benefit from considering more advanced approaches to form classifiers in RVFL networks (see, e.g.,~\cite{ganaie2020minimum, tanveer2021ensemble}).


\section{Conclusion}
\label{sec:conc}

In this paper, we have proposed to improve the efficiency of RVFL networks in distributed classification problems by employing ideas from the HDC/VSA frameworks. 
Notably, we considered a decentralized approach where the  obtained classifiers are able to reach satisfactory results when dealing with local subsets of training data, even without sharing the actual data samples. 
In fact, we choose to employ the RVFL networks to further benefit from its simple design, while retaining the benefits of the fast training and simple RLS or centroids solution.
In the work, we made use of the compression procedure, which uses the binding operation (implemented via the circular convolution) to efficiently implement a distributed classification in which each local agent is able to reach a compelling classification accuracy without overcrowding the network.
The results of the experiments carried out on fully connected networks with varying number of agents, have assessed the performance of the proposed approach with respect to the centralized and local versions. 
Furthermore, we highlighted the results expected in the case of relying on only the local version of a classifier, without sharing any information between the agents.

Certainly, to have a more comprehensive picture of the application of distributed learning theory to the randomized concepts, there are other areas which must be considered. 
Namely, the assumption of simply splitting the local subset should be challenged, while the variegated topologies the network can assume and other different compression techniques should be investigated.
Nonetheless, we believe that this work is able to shine some light on how new paradigms of learning theory can be employed in diverse neural learning schemes, providing a solution to balancing the efficiency/accuracy equilibrium in such methods, which is still an open problem.

\bibliographystyle{IEEEtran} 
\bibliography{references}

\end{document}